\documentclass[wcp]{jmlr}

\usepackage{physics}

\DeclareMathOperator*{\argmin}{arg\,min}

\usepackage{longtable}% for long tables
\usepackage{makecell}

\usepackage{caption}
\usepackage{lineno}

\makeatletter
\let\Ginclude@graphics\@org@Ginclude@graphics 
\makeatother
\title[Adaptation of UBIC for parametric PDE discovery]{Adaptation of uncertainty-penalized Bayesian information criterion for parametric partial differential equation discovery}
\author{\Name{Pongpisit Thanasutives} \Email{thanasutives@ai.sanken.osaka-u.ac.jp}\\
\addr Graduate School of Information Science and Technology, Osaka University, Osaka, Japan
\AND
\Name{Ken{-}ichi Fukui} \Email{fukui@ai.sanken.osaka-u.ac.jp}\\
\addr SANKEN (The Institute of Scientific and Industrial Research), Osaka University,  Osaka, Japan
}

\begin{document}

\maketitle

\begin{abstract}
Data-driven discovery of partial differential equations (PDEs) has emerged as a promising approach for deriving governing physics when domain knowledge about observed data is limited. Despite recent progress, the identification of governing equations and their parametric dependencies using conventional information criteria remains challenging in noisy situations, as the criteria tend to select overly complex PDEs. In this paper, we introduce an extension of the uncertainty-penalized Bayesian information criterion (UBIC), which is adapted to solve parametric PDE discovery problems efficiently without requiring computationally expensive PDE simulations. This extended UBIC uses quantified PDE uncertainty over different temporal or spatial points to prevent overfitting in model selection. The UBIC is computed with data transformation based on power spectral densities to discover the governing parametric PDE that truly captures qualitative features in frequency space with a few significant terms and their parametric dependencies (i.e., the varying PDE coefficients), evaluated with confidence intervals. Numerical experiments on canonical PDEs demonstrate that our extended UBIC can identify the true number of terms and their varying coefficients accurately, even in the presence of noise. The code is available at \url{https://github.com/Pongpisit-Thanasutives/parametric-discovery}.
\end{abstract}


\section{Introduction}
The field of data-driven partial differential equation (PDE) discovery has been advanced progressively with the objective of identifying the governing PDE of a dynamical system automatically. Unlike the traditional derivation of physics from first principles, the data-driven discovery method leverages machine learning techniques applied directly to observed data to achieve greater flexibility and accuracy, as statistical or deep models are tailored to the intricacies of the observed data. A sparse regression based method like SINDy (sparse identification of nonlinear dynamics) \citep{brunton2016discovering, rudy2017data} is typically used to discover nonlinear dynamics, described by a differential equation, via approximating the temporal derivative of a system's state variable using a linear combination of candidate terms. Successful applications of SINDy have been shown in various domains, including but not limited to aerodynamics, biology, and epidemiology \citep{sindy_applications}. 

Throughout the evolution of the field, significant challenges have arisen as primary concerns in the development of data-driven PDE discovery methods. One major challenge that cannot be resolved trivially is determining the optimal regularization hyperparameter(s) for sparse regression. Improper hyperparameter tuning can easily lead to incorrect models, either overfitted or underfitted. To address this issue, PDE discovery methods based on best-subset selection \citep{MIO} have been introduced, offering customizable sparsity in regression models (i.e., the number of nonzero terms or the support size). Nevertheless, these methods sacrifice training speed for greater control over sparsity and the precision of effective candidate terms.

Suppose we have obtained correct support sets that represent potential PDEs of different complexity measures. The model selection step is still required to determine the optimal support size, commonly using an information criterion. In other words, we must decide how many candidate terms should be included in the underlying physics. Conventional criteria like Akaike and Bayesian information criteria (AIC/BIC) \citep{AIC, BIC}, despite being widely used in various scientific disciplines, struggle to select the optimal support size for the governing equation, favoring overly complex models, when no PDE simulation is performed before the AIC calculation \citep{nPIML}. In contrast to SINDy-AIC \citep{SINDy-AIC}, which necessitates simulations of multiple potential PDEs before calculating AIC scores, we are more interested in achieving reliable and fast model selection without requiring any PDE solutions. As discussed by \cite{SGTR} and \cite{improvedAIC}, minimizing AIC with a variable threshold benchmark or tolerance can lead to selecting an incorrect governing PDE. Setting the threshold too low results in an overly sparse PDE, while setting it too high leads to an overfitted PDE. Following Ockham's razor, the uncertainty-penalized information criterion (UBIC) \citep{UBIC, thanasutives2024proof} has been recently proposed to balance model accuracy and complexity while accounting for PDE uncertainty. UBIC incorporates the quantified uncertainty of PDE coefficients (model parameters) to avoid overfitting. The PDE uncertainty is achieved by employing Bayesian regression to infer posterior distributions of the coefficients, from which the mean and covariance are computed. The resulting coefficient of variation penalizes the base BIC adaptively to ensure the selection of a parsimonious and stable equation as the governing PDE. Different from UQ-SINDy \citep{UQ-SINDy}, which quantifies the uncertainty of PDE coefficients (sparsified by priors) but does not make use of this uncertainty to impact model selection, UBIC employs its quantified uncertainty to address the overfitting problem. 

The central challenge of this paper is to find the number of spatially or temporally varying coefficients in the governing parametric equation. In cases where the PDE coefficients are not just constant, the governing PDE identification becomes much more difficult. The aforementioned methods were originally proposed to tackle the PDE discovery problem with constant coefficients but are unable to handle varying coefficients. 

To discover parametric PDEs, \cite{SGTR} proposed sequential grouped threshold ridge regression (SGTR), where a separate sparse regression is solved for each time step (or spatial point for the case of spatially dependent PDEs) to attain corresponding time-varying coefficients. Although this method is able to obtain coefficients for each specific point, it cannot provide a symbolic expression for varying coefficients. SGTR relies on AICc (AIC with a correction for small sample sizes) \citep{AICc} losses, which do not promote parsimonious PDEs and thus probably mislead us into selecting unnecessarily complex PDEs. Moreover, the SGTR algorithm is noise-intolerant, which may result in discovering an incorrect PDE form or structure. We refer readers to \cite{robustSGTR} for an extended SGTR with double low-rank decompositions (to denoise sparse outlying entries) and to \cite{luo2023physics} for a kernelized SGTR that improves noise robustness, albeit at the increased computational cost. Although these methods have shown noise robustness, their model selection relies on either tweaking the amount of norm-based regularization applied or using AIC with costly PDE simulations. \cite{DLGA-PDE} tackled parametric PDE discovery problems using DLGA-PDE, a genetic algorithm-based framework that employs deep learning as a mesh-free function approximator. Although the framework can discover the governing parametric PDE in closed form, giving symbolic expressions for varying coefficients, the PDE structure identification step is computationally intensive because of the need to train neural networks repeatedly on different PDEs found during the genetic algorithm's learning process. Furthermore, tuning the regularization hyperparameters in the fitness function is problematic because it depends on (biased) human experience, making the identification process difficult to automate, as discussed earlier. 

Our main contribution is the new extension of UBIC, adapted to solve parametric PDE discovery problems. This extended UBIC leverages quantified PDE uncertainty as a complexity penalty to address the overfitting issue in the model selection step. It also inherits the benefits of UBIC, including no computationally expensive PDE simulation required (low computational resource requirements) and minimal dependence on hyperparameter tuning (\citealp[see][]{UBIC}). We find that using the UBIC score of the power spectral densities of the regression model and the temporal derivative effectively prevents the selection of overfitted PDEs, promoting the parsimonious model that truly captures qualitative features in frequency space. As a byproduct of computing the UBIC, we provide confidence intervals for all coefficients, each evaluated at a particular time step or spatial grid point. 

\section{Methodology}
\subsection{Problem Formulation}
Let us consider the following parametric form of governing partial differential equations: 

\begin{equation} \label{eq:problem_formulation}
	u_{t} = \mathcal{N}(u, u_{x}, u_{xx}, \dots; \mu(x, t)) = \sum_{j=1}\mathcal{N}_{j}(u, u_{x}, u_{xx}, \dots)\mu_{j}(x, t).
\end{equation}

\noindent We aim to identify the nonlinear operator $\mathcal{N}$, which involves spatial derivatives of the state variable $u$, whose realization $\vec{\mathrm{U}} \in \mathbb{R}^{N_{x} \times N_{t}}$ is given in a spatio-temporal grid. $\mathcal{N}$ is parameterized by $\mu(x, t)$, reducible to either $\mu(x)$ or $\mu(t)$---spatially or temporally varying functions. 

\subsection{Sparse Regression}
Suppose, without loss of generality to spatially varying cases, Equation \eqref{eq:problem_formulation} can be formulated as systems of linear equations, with temporal dependency. Given there are $N_{t}$ time steps and $N_{x}$ spatial points, the system evaluated at a time $t_{i}$ is expressed by 

\begin{equation}
	\vec{\mathrm{U}^{i}_{t}} = \mathbf{Q}^{\boldsymbol{i}}\vec{\xi^{i}} = \sum^{N_{q}}_{j=1}\xi^{i}_{j}\vec{q^{i}_{j}};\, \mathbf{Q}^{\boldsymbol{i}} = \begin{pmatrix} 
	\vline & \vline & \vline & \vline \\
	\vec{q^{i}_{1}} & \cdots & \vec{q^{i}_{j}} & \cdots \\
	\vline & \vline & \vline & \vline
\end{pmatrix} \in \mathbb{R}^{N_{x} \times N_{q}}.
\end{equation}

\noindent $\vec{\mathrm{U}_t}$ is the time derivative numerically computed with Kalman smoothing. Every $\mathbf{Q}^{\boldsymbol{i}}$ comprises the identical $N_{q}$ candidate terms, which are presumed overcomplete, with each term possibly appearing in the true $\mathcal{N}$. We define the candidate library $\mathbf{Q}$ as a block-diagonal matrix constructed by all $\mathbf{Q}^{\boldsymbol{i}}$ matrices, deriving a single system for the parametric PDE discovery problem: $\vec{\mathrm{U}_{t}} = \mathbf{Q}\vec{\Xi}$. We achieve the best-subset solution with $s_{k}$ support size via solving the sparse regression separately for each time step: 

\begin{equation} \label{eq:sparse_regression}
	\vec{\hat{\Xi}} = \min_{\vec{\Xi}}\sum^{N_{t}}_{i=1}\norm{\vec{\underline{\mathrm{U}}^{i}_{t}} - \mathbf{\underline{Q}}^{\boldsymbol{i}}\vec{\xi^{i}}}_{2}^{2} + \lambda\norm{\vec{\xi^{i}}}^{2}_{2},\, \textrm{such that}\, \norm{\vec{\xi^{i}}}_{0} = s_{k}, \forall i \leq N_{t}; 
\end{equation}

\noindent where $\vec{\underline{\mathrm{U}}}$ and $\mathbf{\underline{Q}}$ are the validation data on which $\vec{\Xi} \in \mathbb{R}^{N_{q} \times N_{t}}$ is evaluated. The best-subset solver based on mixed-integer optimization (MIOSR) \citep{MIO} is used to gather $\vec{\xi^{i}}$ of consecutive support sizes with $\lambda = 0$. Note that MIOSR is preferred here over SGTR to ensure that potential PDEs with certain support sizes are not overlooked. We can impose the group sparsity in $\vec{\Xi}$ by controlling that the support set $\{j \mid \abs{\xi^{i}_{j}} > 0\}, \forall i \leq N_{t}$ is the same for every time step. Remark that the group sparsity condition is not necessary before optimizing for the optimal number of nonzero terms (support size) $s^{*} \in \{s_{k} \mid k = 1, 2, \dots\}$. Because we cannot infer $s^{*}$ solely from Equation \eqref{eq:sparse_regression}, and the model selection step is performed next. 

\subsection{Model Selection}
We minimize an information criterion to select the optimal support size $s^{*}$. An information criterion is usually expressed by 

\begin{equation} \label{eq:information_criterion}
	-2\log L(\vec{\hat{\Xi}}) + \mathcal{C}(a_{N}, \vec{\hat{\Xi}}, \mathcal{P}); 
\end{equation}

\noindent where $L$ is the likelihood function, and $\mathcal{C}(a_{N}, \vec{\hat{\Xi}}, \mathcal{P})$ is the complexity penalty corresponding to the criterion. For example, $2s_{k}$ and $\log(N)s_{k}$; where $N = N_{x}N_{t}$, is the complexity penalty for AIC and BIC, respectively. We regard  $\mathcal{P}$ as any other necessary information, e.g., the complexity measures of ICOMP (informational complexity criterion) \citep{ICOMP} or the UBIC's quantified PDE uncertainty. Considering a particular support size $s_{k}$, the extension of the original UBIC (incorporating a fixed threshold $\zeta = 10^{-5}$ to prevent underflowing) is proposed for parametric PDE discovery as follows: 

\begin{equation} \label{eq:ubic}
\begin{split}
	\mathrm{UBIC}_{k} &= N\log(\frac{2\pi \norm{\vec{\mathrm{U}_{t}} - \mathbf{Q}\vec{\hat{\Xi}}}^{2}_{2}}{N}+\zeta) + \log(N)(\mathfrak{U} + s_{k});\\
	\mathfrak{U} &= 10^{\lambda^{*}}\bar{\mathrm{V}},\, \bar{\mathrm{V}} = \frac{\mathrm{V}}{\mathrm{V}_{\max}},\, \mathrm{V} = \Sigma^{N_{t}}_{i=1}R_{i},\, \textrm{and} \,R_{i} = \frac{\Sigma^{s}_{j=1}\sigma^{i}_{j}}{\norm{\vec{\hat{\xi}^{i}}}_{1}}.
\end{split}
\end{equation}

\noindent According to \cite{UBIC}, we compute the uncertainty $\mathfrak{U}$ (of the $s_{k}$-support-size PDE) using the tuned data-dependent $\lambda^{*}$ and the scaled coefficient of variation $\bar{\mathrm{V}}$. At each time step, $\mathrm{V}$ accumulates an instability ratio $R_{i}$ of the total posterior standard deviation and the coefficient vector L1-norm, both obtained by applying Bayesian automatic relevance determination (ARD) regression \citep{ARD}. $\mathrm{V}_{\max}$ is the maximum value of $\mathrm{V}$ over all available support sizes. Essentially with the temporal (or spatial) accumulation, we can extend the original UBIC for parametric PDE discovery. Remark that if we weaken the group sparsity assumption, we allow each $\vec{\hat{\xi}^{i}}$ to have a unique support set---$S^{i} = \{j \mid \hat{\xi}^{i}_{j} \neq 0\}$ with an equal support size (cardinality) $\abs{S^{i}} = s_{k}$, leading to a reasonable increase in the complexity penalty simply by considering $s_{k} \times \abs{\{S^{i}\}}$ instead of just having a $s_{k}$ fixed across the temporal or spatial axis. 

\textbf{Power spectral density based transformation}. The validation data  $\mathbf{\underline{Q}}$ in frequency space is obtained by applying discrete Fourier transformation along the temporal axis to every 

\begin{equation*}
\mathbf{Q}_{\boldsymbol{j}} = 
\begin{pmatrix} 
	\vline & \vline & \vline & \vline \\
	\vec{q^{1}_{j}} & \cdots & \vec{q^{i}_{j}} & \cdots \\
	\vline & \vline & \vline & \vline
\end{pmatrix} \in \mathbb{R}^{N_{x} \times N_{t}}, 
\end{equation*}

\noindent and removing entries that correspond to low-power frequencies---less than the ninety percentile. The transformation technique is not only beneficial when deciding the optimal coefficient vector with $s_{k}$ support size, but also when selecting the optimal $s^{*}$. More specifically, we generalize the residual sum of squares (RSS) to $\norm{T(\vec{\mathrm{U}_{t}}) - T(\mathbf{Q}\vec{\hat{\Xi}})}^{2}_{2}$; where $T$ denotes a function, which transforms $\vec{\mathrm{U}_{t}}$ and $\mathbf{Q}\vec{\hat{\Xi}}$ to new representations, i.e., mapping $T(\vec{\mathrm{U}_{t}}) = \vec{\tilde{\mathrm{U}}_{t}}$. In this work, every $\vec{\tilde{\mathrm{U}}^{i}_{t}}$ is a numerical result from a trapezoidal integration applied along the spatial axis (the frequency/temporal axis for a spatially dependent PDE) of estimated power spectral density (PSD) using a periodogram (see \texttt{scipy.integrate.trapezoid} and \texttt{scipy.signal.periodogram}). The PSD representation is a good choice because of its clear characteristics (see Appendix \ref{app:psd}), which exhibits larger values for true data-generating frequencies \citep{PSD}. The integration trick is to limit the sample number, and therefore facilitates the model selection step, as it is known that conventional information criteria tend to select overfitted PDEs when the number of samples is large \citep{nPIML, UBIC}. The integration is ablated if the estimated PSD is already a one-dimensional vector. 

\textbf{Support size filtering}. When tuning for $\lambda^{*}$ to ultimately achieve $s^{*}$, we only consider a sequence of significant support sizes $[k \mid \mathrm{BIC}_{k} < \mathrm{BIC}_{k^{\prime}} \land p_{\textrm{MW}}([\mathrm{BIC}^{i}_{k}], [\mathrm{BIC}^{i}_{k^{\prime}}]) < \textrm{p}_{\textrm{cut}}]$; where $k^{\prime} = \argmin_{k^{\prime} < k}\textrm{BIC}_{k}$. We refer to $[\mathrm{BIC}^{i}_{k}]$ as a collection of temporally varying BIC scores, each calculated on $T(\vec{\mathrm{U}^{i}_{t}})$ and $T(\mathbf{Q}^{\boldsymbol{i}}\vec{\hat{\xi}^{i}})$. $p_{\textrm{MW}}([\mathrm{BIC}^{i}_{k}], [\mathrm{BIC}^{i}_{k^{\prime}}])$ is the p-value resulted from the Mann–Whitney test with the alternative hypothesis stating that the distribution underlying $[\mathrm{BIC}^{i}_{k}]$ is stochastically less than the distribution underlying $[\mathrm{BIC}^{i}_{k^{\prime}}]$, hence the significant improvement by increasing the support size from $s_{k^{\prime}}$ to $s_{k}$. The median of $\{\textrm{p} \mid \textrm{p} = p_{\textrm{MW}}([\mathrm{BIC}^{i}_{k+1}], [\mathrm{BIC}^{i}_{k}]) \land \textrm{p} < 0.01\}$ defines $\textrm{p}_{\textrm{cut}}$, ensuring that we consider only the top improvements with at least $99$\% confidence. We recommend researchers explore a generalized percentile-based notion beyond sticking to the strict median. 

\begin{figure}[t]
\centering
\includegraphics[width=0.8\textwidth]{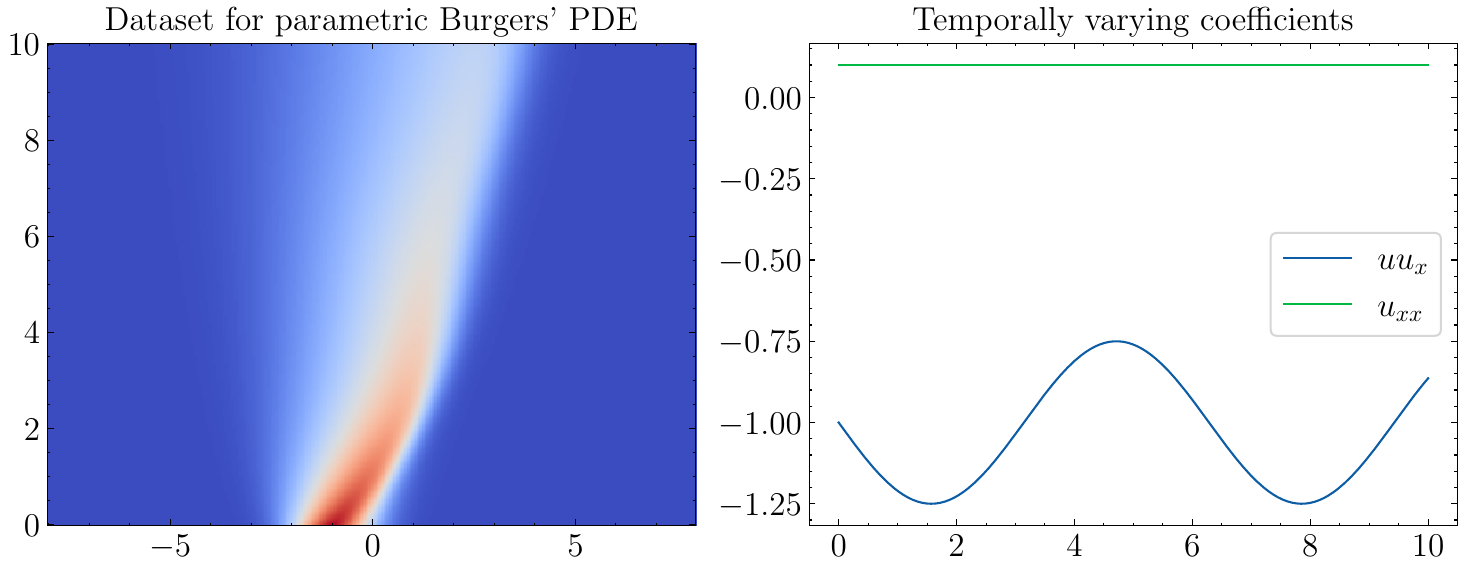}
\caption{Visualization of the state variable and the temporally varying coefficients for the parametric Burgers' PDE dataset.}
\label{fig:burgers_dataset}
\end{figure}

\begin{figure}[t]
\centering
\begin{subfigure}
	\centering
	\includegraphics[width=0.8\textwidth]{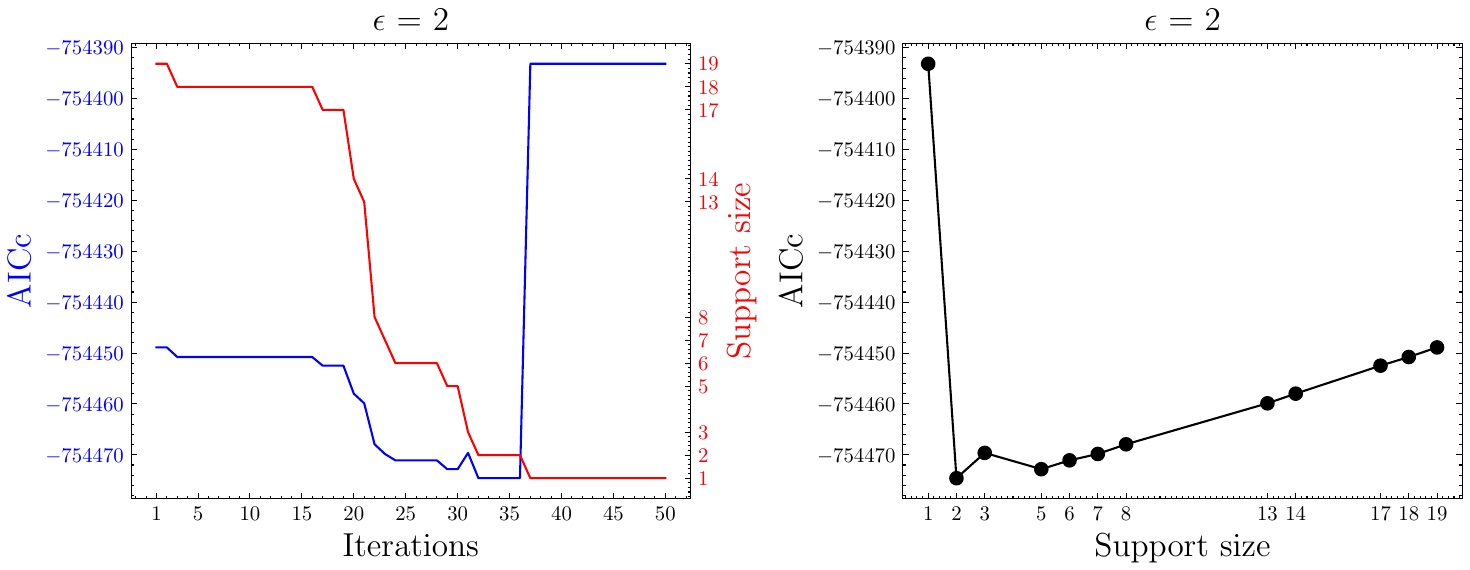}
\end{subfigure}
\begin{subfigure}
	\centering
	\includegraphics[width=0.8\textwidth]{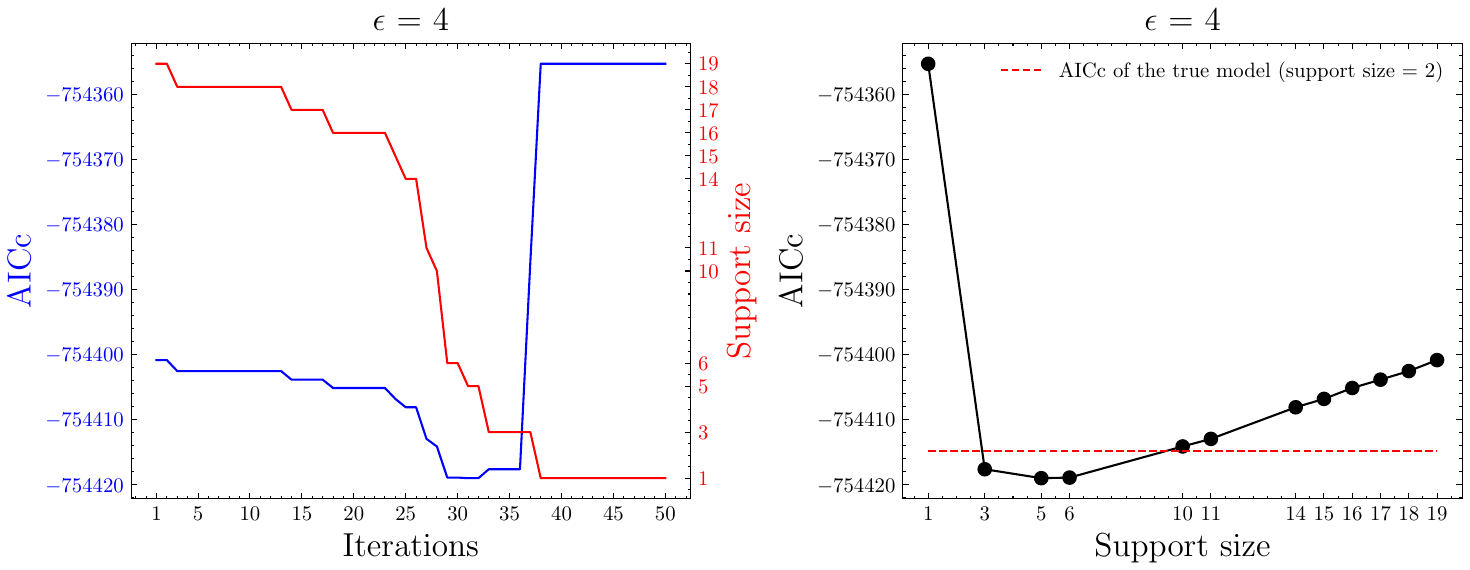}
	\end{subfigure}
\caption{\textbf{Parametric Burgers' PDE}: Model selection using AICc employed by the SGTR algorithm under noisy situations. According to \cite{SGTR}, RSS is calculated based on every L2-normalized $\vec{\mathrm{U}^{i}_{t}}$ and $\mathbf{Q}^{\boldsymbol{i}}$.}
\label{fig:burgers_sgtr_noise}
\end{figure}

\begin{figure}[t]
\centering
\begin{subfigure}
	\centering
	\includegraphics[width=0.8\textwidth]{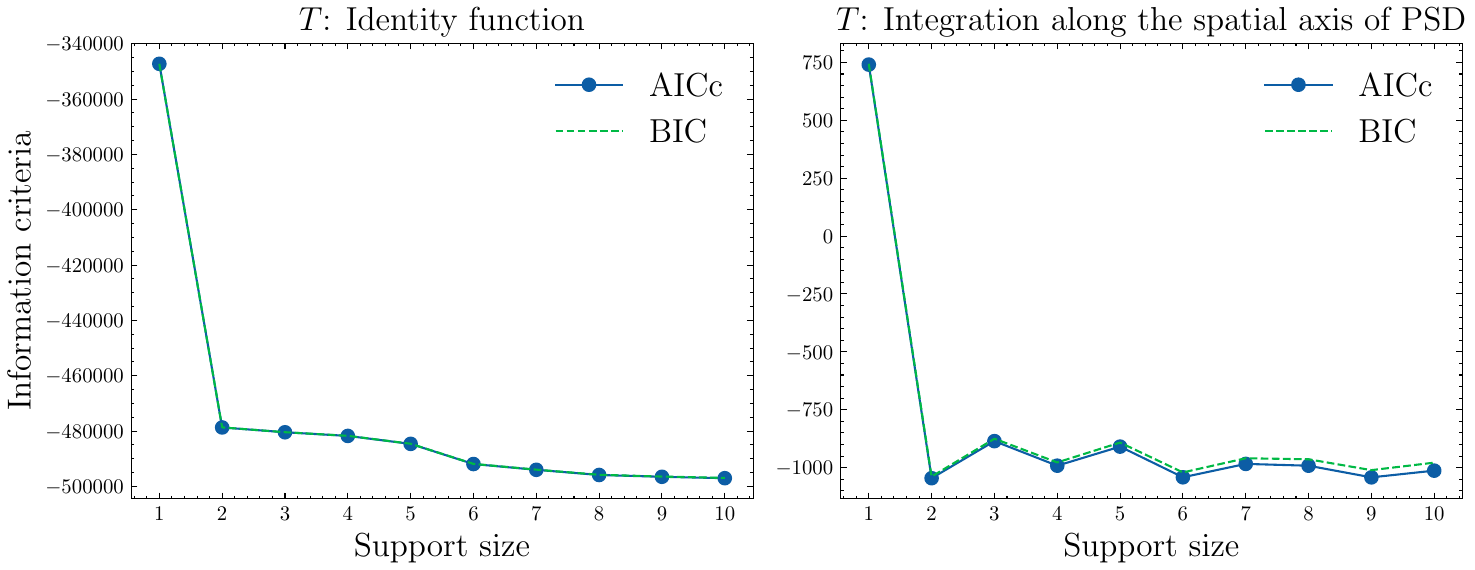}
	\caption*{$\epsilon = 2$}
\end{subfigure}
\begin{subfigure}
	\centering
	\includegraphics[width=0.8\textwidth]{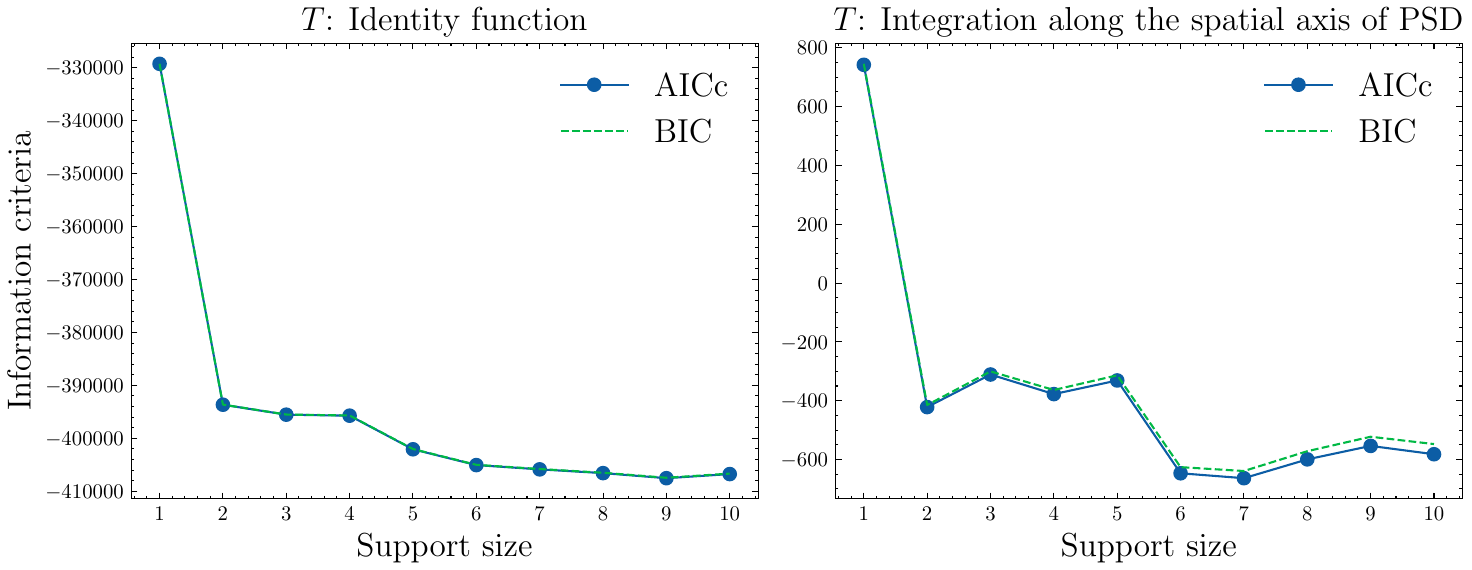}
	\caption*{$\epsilon = 4$}
	\end{subfigure}
\caption{\textbf{Parametric Burgers' PDE}: Information criteria are calculated with different transformations $T$. Potential best subsets are obtained through MIOSR (not SGTR) and then validated using Equations \eqref{eq:sparse_regression}. We use unnormalized $\vec{\mathrm{U}^{i}_{t}}$ and $\mathbf{Q}^{\boldsymbol{i}}$ when calculating RSS.}
\label{fig:burgers_aicbic_noise}
\end{figure}

\begin{figure}[t]
\centering
\begin{subfigure}
	\centering
	\includegraphics[width=0.4\textwidth]{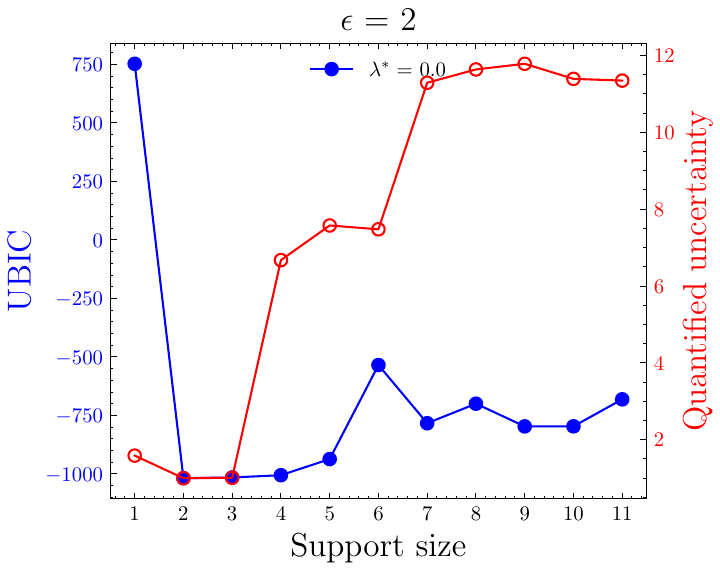}
	\includegraphics[width=0.4\textwidth]{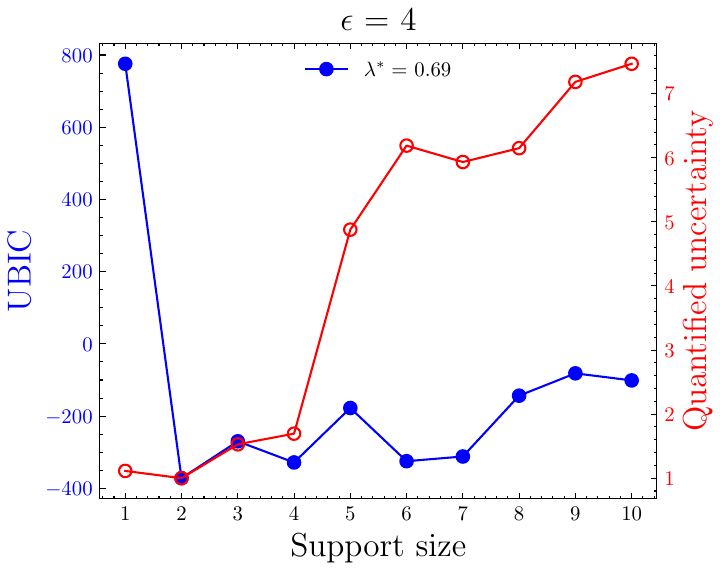}
\end{subfigure}
\begin{subfigure}
	\centering
	\includegraphics[width=0.4\textwidth]{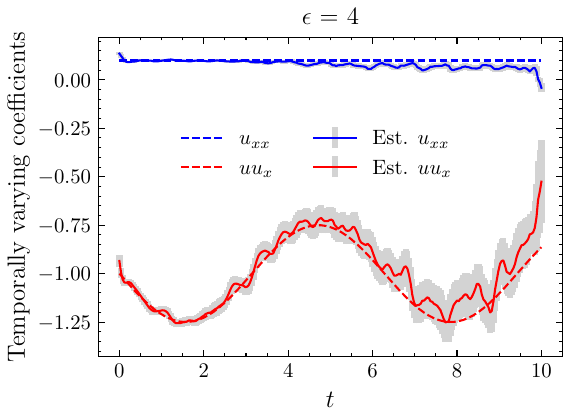}
	\includegraphics[width=0.4\textwidth]{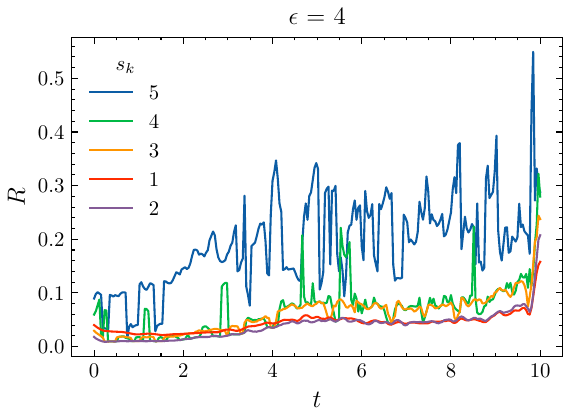}
\end{subfigure}
\caption{\textbf{Parametric Burgers' PDE}: (a) Model selection using our proposed UBIC with the PSD-based transformation under noisy situations. (b) Confidence intervals for the temporally varying coefficients and the instability (over time) of the potential PDEs with different support sizes.}
\label{fig:burgers_ubic_noise}
\end{figure}

\section{Numerical Experiments}
The code for reproducing our results is available at \url{https://github.com/Pongpisit-Thanasutives/parametric-discovery}.
\subsection{Parametric Burgers' PDE with Diffusive Regularization}
The equation is defined with the constant diffusion of $0.1$ and the time-varying coefficient $a(t)$ for the nonlinear advection. 

\begin{equation}\label{eq:burgers}
u_t = a(t) uu_x + 0.1 u_{xx};\, a(t) = -(1 + \frac{\sin(t)}{4}),\, x \in [-8, 8],\, \textrm{and} \, t \in [0, 1].
\end{equation}

\noindent The time-dependent Burgers’ equation with periodic boundary conditions is numerically solved using a spectral method on a discretized spatio-temporal grid of size $N_{x} \times N_{t} = 256 \times 256$. The noise-free solution $\vec{\mathrm{U}}$ (shown in the left of Figure \ref{fig:burgers_dataset}) is perturbed with $\epsilon$\%-sd (standard deviation) Gaussian noise given by $\epsilon \times \mathrm{sd}(\vec{\mathrm{U}}) \times \mathcal{N}(0, 1)$ to generate a noisy situation. Generally, in this work, we apply a Savitzky-Golay filter to smooth the resulting distorted data before derivative computations. The candidate terms include powers of $u$ up to cubic order, which are multiplied by spatial derivatives of $u$ up to fourth order. For this canonical example, PDE discovery methods are tested with $\epsilon = 2, 4$.

We investigate the most common method for discovering parametric PDEs, namely SGTR, and then inspect some of its problems and disadvantages. As seen in Figure \ref{fig:burgers_sgtr_noise}, in the case where the noise level is lower, $\epsilon = 2$, SGTR converges and identifies the true governing PDE, however, some support sizes are left unexplored, raising a concern regarding the algorithm design based on fitting Ridge solver with hard thresholding to remove small-norm coefficients iteratively. The fact that the solutions of different support sizes are not guaranteed to be optimal may have fortuitously led to the finding of the true support size. The concern turns into a real problem in the higher noise case, $\epsilon = 4$. Because the actual support size has never been checked by the SGTR algorithm, the false PDE is identified inevitably. Moreover, the model selection guided by the AICc loss has led us to an overfitted model anyway, as the number of training samples is large. 

Considering a sequence of models with consecutive support size (no skip) obtained through MIOSR to satisfy Equation \eqref{eq:sparse_regression}, we test whether our transformation $T$ on $\vec{\mathrm{U}_{t}}$ and $\mathbf{Q}\vec{\hat{\Xi}}$ alleviate the overfitting problem. As seen in Figure \ref{fig:burgers_aicbic_noise}, when $\epsilon = 2$, our PSD-based transformation is able to prevent selecting overfitted models and help us identify the governing model, while the beneficial effect is attenuated due to the higher noise level of $\epsilon = 4$. It is noteworthy that without any transformation ($T$ is an identity function), we cannot at all prevent the overfitting problem. 

In Figure \ref{fig:burgers_ubic_noise}(a), we test our proposed UBIC with the PSD-based transformation in identifying the governing equation. Under both noisy levels, the quantified uncertainty $\mathfrak{U}$ is able to determine the true support size---$2$ for this canonical PDE. Consequently, it is natural that the UBIC utilizes the quantified uncertainty to penalize overfitted models and successfully identify the correct form of the governing Burgers' equation despite the high noise level. In Figure \ref{fig:burgers_ubic_noise}(b), we plot the posterior coefficient and (double) standard deviation derived from Bayesian ARD regression, illustrating the 95\% confidence interval. The confidence interval pinpoints when noticeable instability in the estimated posterior coefficient increases. In this case, a common behavior observed over different support sizes is the increasing instability over time steps, indicating the challenge in estimating varying coefficients (and also uncovering their symbolic expressions in Appendix \ref{app:symbolic_discovery}) over an extended time period. The observation suggests that long-range approximations should be circumvented to reveal the overall pattern of the varying coefficients accurately and maintain bounded instability. In Appendix A, we also employ symbolic regression to accurately recover the varying coefficients in closed form for each PDE dataset we experiment with. 

\begin{figure}[t]
\centering
\includegraphics[width=0.8\textwidth]{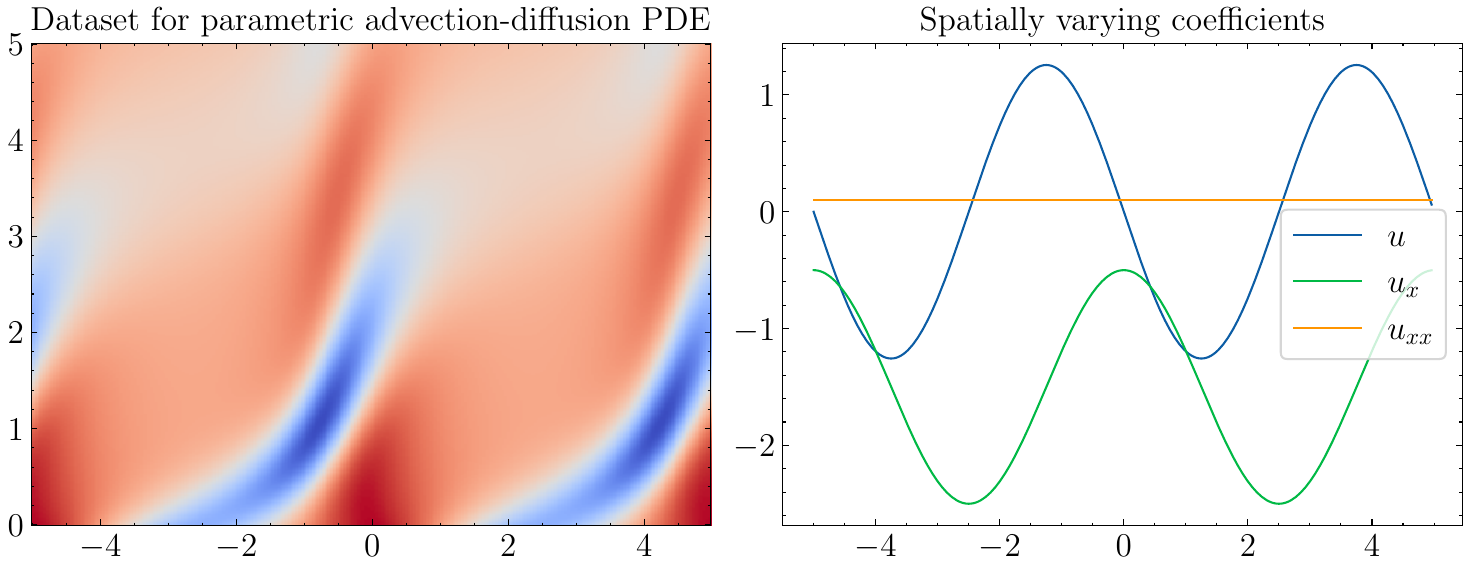}
\caption{Visualization of the state variable and the spatially varying coefficients for the parametric advection-diffusion PDE dataset.}
\label{fig:ad_dataset}
\end{figure}

\begin{figure}[t]
\centering
\begin{subfigure}
	\centering
	\includegraphics[width=0.8\textwidth]{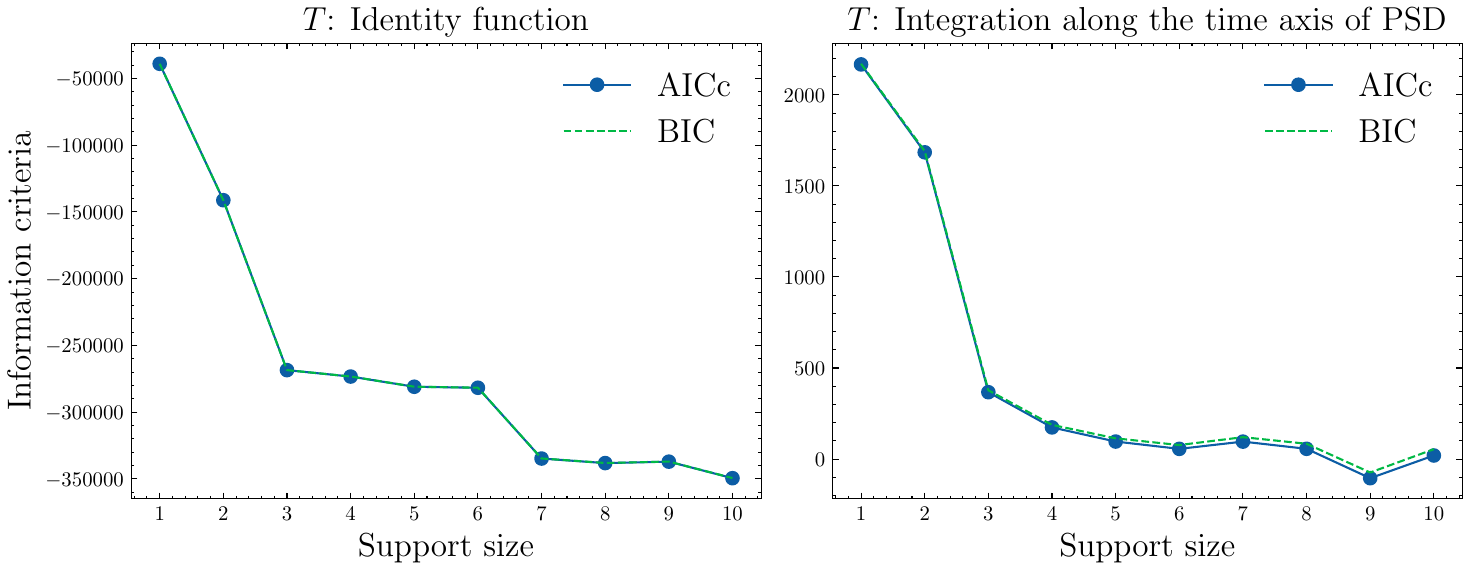}
	\caption*{$\epsilon = 2$}
\end{subfigure}
\begin{subfigure}
	\centering
	\includegraphics[width=0.8\textwidth]{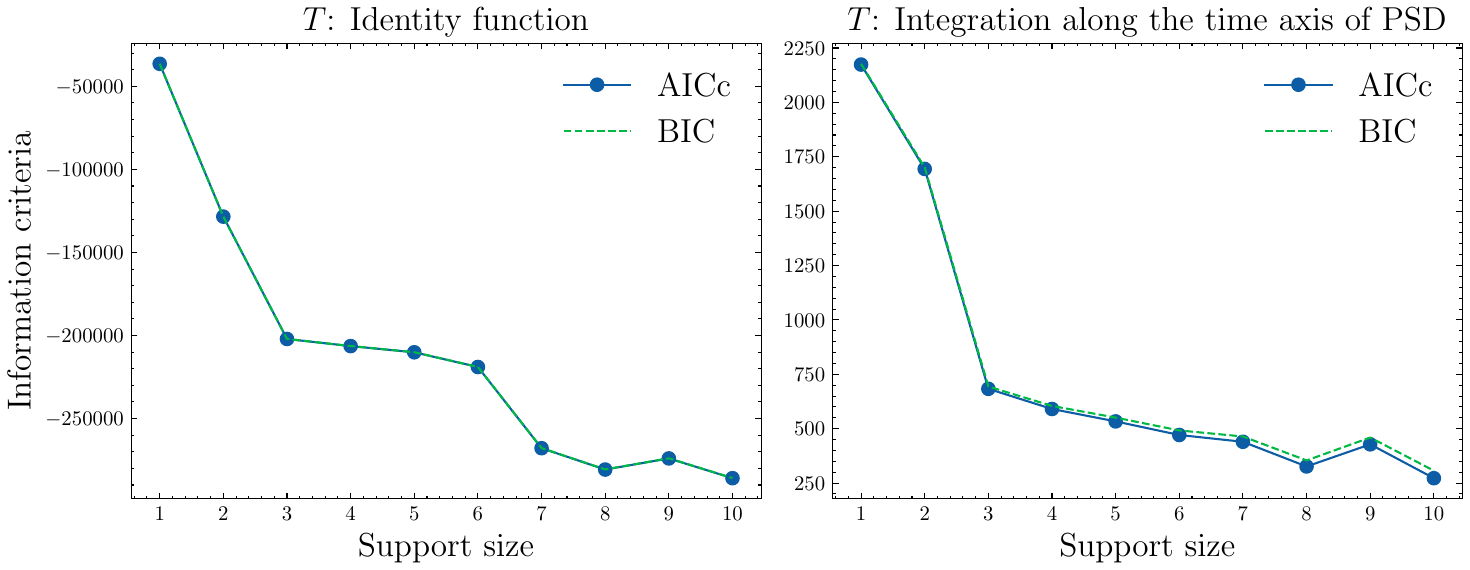}
	\caption*{$\epsilon = 4$}
	\end{subfigure}
\caption{\textbf{Parametric advection-diffusion PDE}: Information criteria are calculated with different transformations $T$. Potential best subsets are obtained through MIOSR and then validated using Equations \eqref{eq:sparse_regression}. We use unnormalized $\vec{\mathrm{U}^{i}_{t}}$ and $\mathbf{Q}^{\boldsymbol{i}}$ when calculating RSS.}
\label{fig:ad_aicbic_noise}
\end{figure}

\begin{figure}[t]
\centering
\begin{subfigure}
	\centering
	\includegraphics[width=0.4\textwidth]{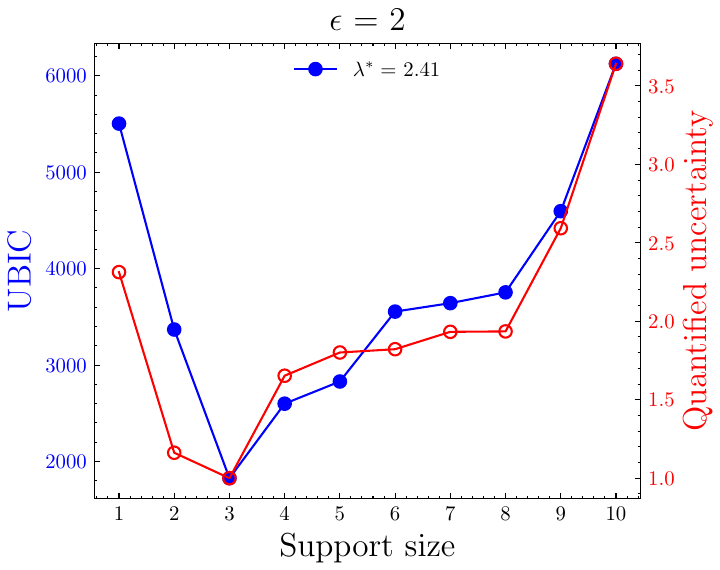}
	\includegraphics[width=0.4\textwidth]{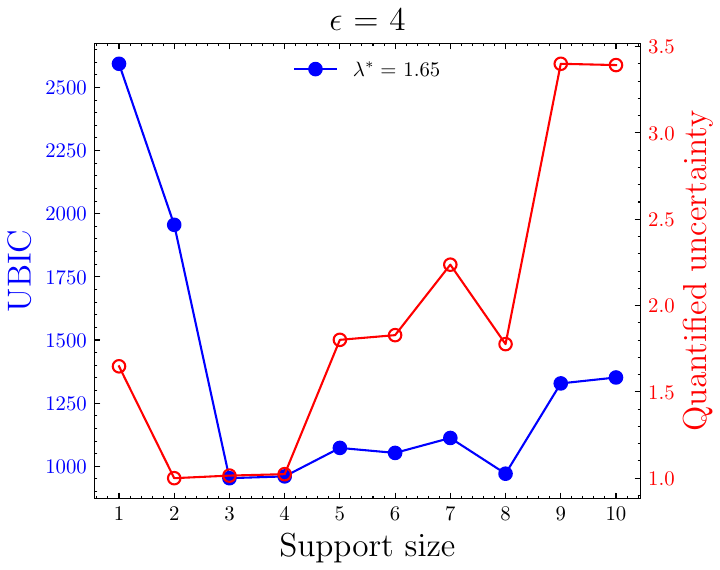}
\end{subfigure}
\begin{subfigure}
	\centering
	\includegraphics[width=0.4\textwidth]{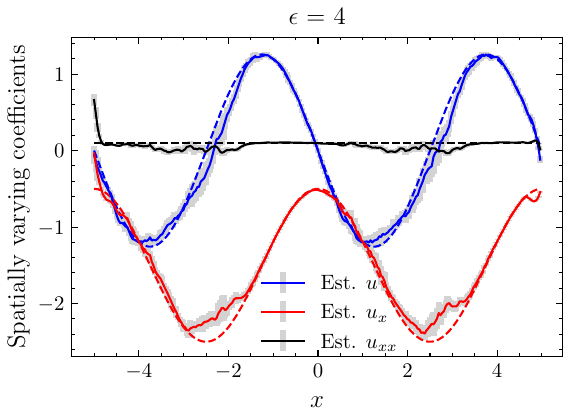}
	\includegraphics[width=0.4\textwidth]{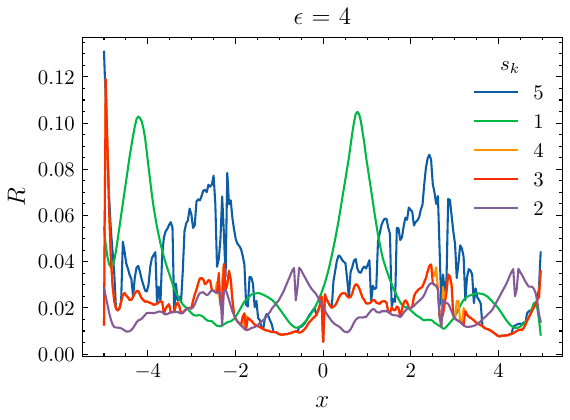}
\end{subfigure}
\caption{\textbf{Parametric advection-diffusion PDE}: (a) Model selection using our proposed UBIC with the PSD-based transformation under noisy situations. (b) Confidence intervals for the spatially varying coefficients and the instability (over spatial points) of the potential PDEs with different support sizes. Dotted lines (\texttt{---}) denote the true varying coefficients.}
\label{fig:ad_ubic_noise}
\end{figure}

\subsection{Spatially Dependent Advection-Diffusion (AD) PDE}
The advection-diffusion PDE describes the transport of a physical quantity in a varying velocity field with diffusion. We experiment with the following equation with a spatially dependent velocity: 

\begin{equation} \label{eq:advection}
u_t = c'(x)u + c(x)u_x + 0.1u_{xx};\, c(x) = -1.5 + \cos(\frac{2\pi x}{5}),\ x \in [-5, 5],\, \textrm{and} \, t \in [0, 5].
\end{equation}

\noindent We use a spectral method to numerically solve the PDE on a periodic domain of size $N_{x} \times N_{t} = 256 \times 256$. The clean solution is depicted in Figure \ref{fig:ad_dataset}. The candidate library consists of powers of $u$ up to cubic, which are multiplied by spatial derivatives of $u$ up to fourth order. For this PDE we also experiment with the two noise levels: $\epsilon = 2$ and $\epsilon = 4$. 

Considering the $\epsilon = 2$ case in Figure S1 (`S' refers to a figure in the Supplementary material), the SGTR algorithm can discover the governing equation, possibly because of the suboptimal ridge solutions being compared. However, in the higher noise case, $\epsilon = 4$, the overfitted PDE with $4$ nonzero terms is chosen instead, as favored by the AICc loss. 

As seen in Figure \ref{fig:ad_aicbic_noise}, although the overfitting problem is not completely solved, it is partially mitigated because the inclination to choose overly complicated PDEs diminishes after applying the PSD-based transformation. This observation aligns with the fact that, in the lower noise case, $\epsilon = 2$, the SGTR algorithm would discover an overfitted PDE instead if the L2-normalization has not been done on $\vec{\mathrm{U}^{i}_{t}}$ and $\mathbf{Q}^{\boldsymbol{i}}$. 

We quantify the PDE uncertainty of potential parametric models whose support size is up to $10$. For the $\epsilon = 2$ case in Figure \ref{fig:ad_ubic_noise}(a), the quantified uncertainty points exactly to the PDE with the correct form and support size, enabling us to determine the governing equation using the UBIC. For the $\epsilon = 4$ case, although the $3$-support-size parametric model exhibits the second minimum uncertainty, it is still useful for the UBIC to identify the parametric model as the governing equation because it outperforms the $2$-support-size in terms of RSS. We do not find any sufficient significance in adding more candidate terms. 

In Figure \ref{fig:ad_ubic_noise}(b), the posterior coefficient and standard deviation are plotted to illustrate the 95\% confidence interval. The instability of the varying coefficients is revealed particularly near the left spatial boundary and dynamically changing locations. This observation can help refine the quantified PDE uncertainty by avoiding these dynamically changing locations. For example, considering the spatial locations starting from $x = -5 + 4\Delta x = -4.84375$ only, the refined PDE uncertainty of the $3$-support-size parametric model becomes the minimum value. 

\begin{figure}[t]
\centering
\includegraphics[width=0.8\textwidth]{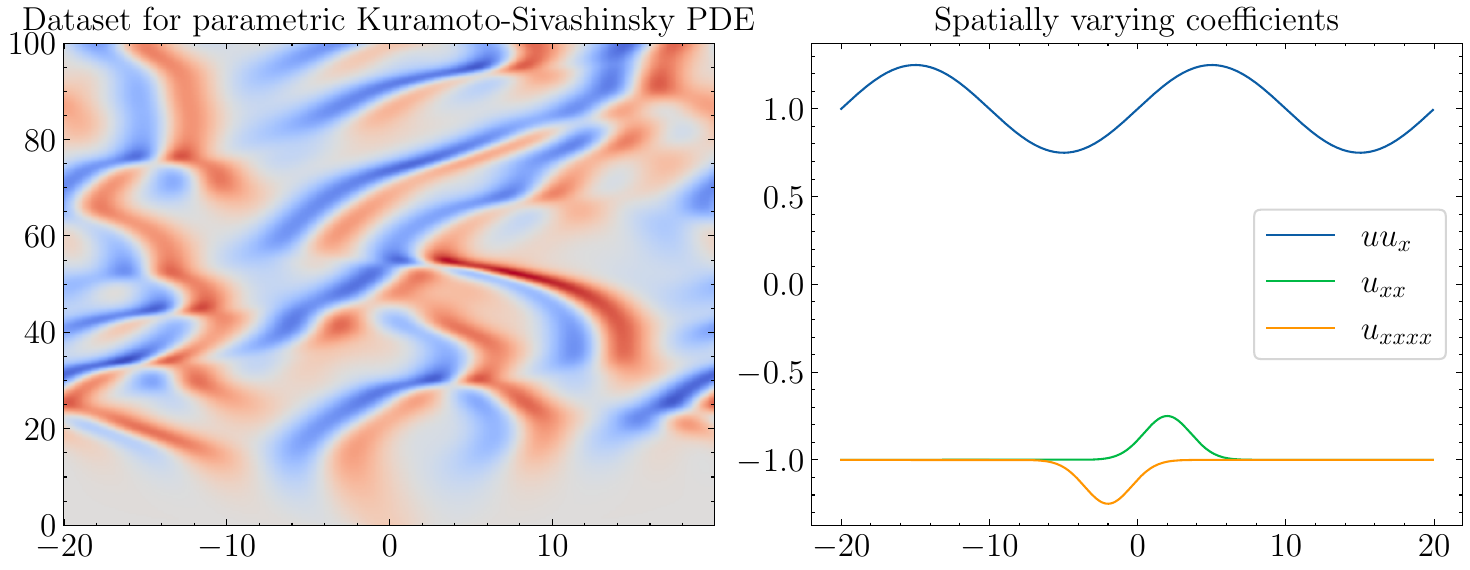}
\caption{Visualization of the state variable and the spatially varying coefficients for the parametric Kuramoto-Sivashinsky PDE dataset.}
\label{fig:ks_dataset}
\end{figure}

\begin{figure}[t]
\centering
\begin{subfigure}
	\centering
	\includegraphics[width=0.8\textwidth]{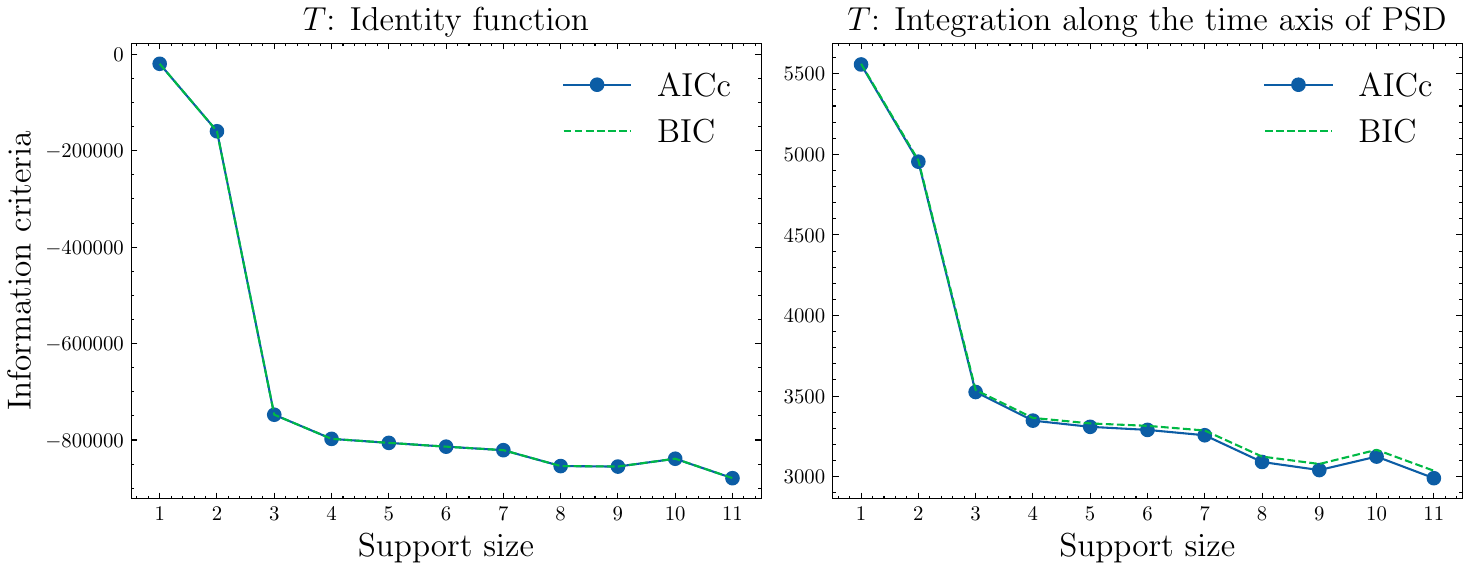}
\end{subfigure}
\begin{subfigure}
	\centering
	\includegraphics[width=0.8\textwidth]{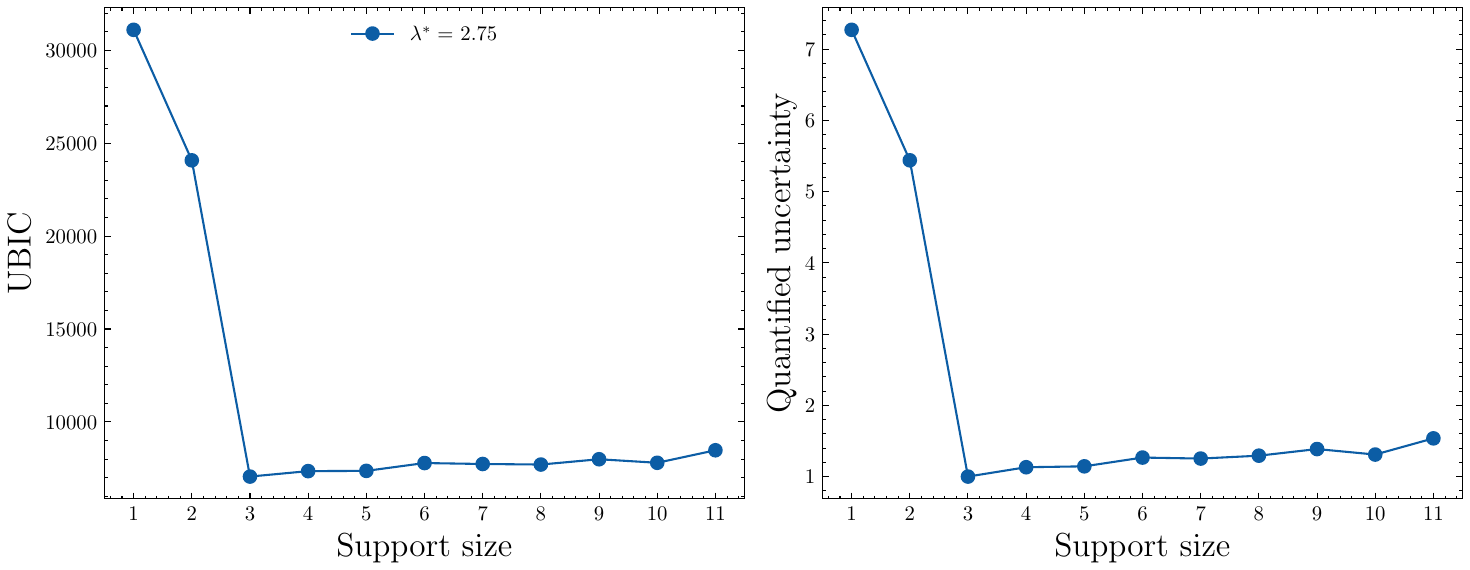}
\end{subfigure}
\begin{subfigure}
	\centering
	\includegraphics[width=0.4\textwidth]{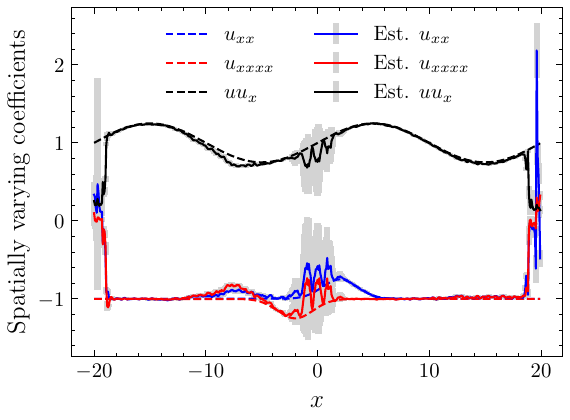}
	\includegraphics[width=0.4\textwidth]{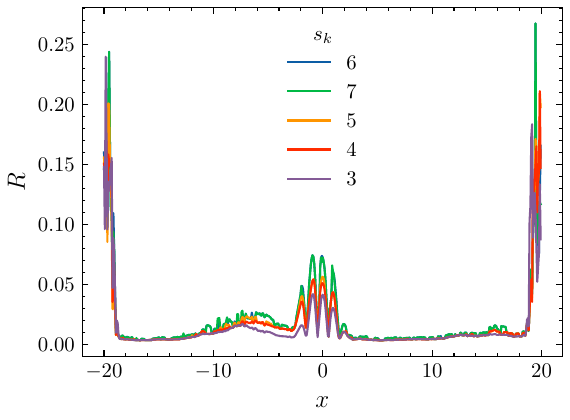}
\end{subfigure}
\caption{\textbf{Parametric Kuramoto-Sivashinsky PDE.} (a) Under the noisy situation, where we set $\epsilon = 0.01$, conventional information criteria are calculated with different transformations $T$. Potential best subsets are obtained through MIOSR and then validated using Equations \eqref{eq:sparse_regression}. We use unnormalized $\vec{\mathrm{U}^{i}_{t}}$ and $\mathbf{Q}^{\boldsymbol{i}}$ when calculating RSS. (b) Model selection using our proposed UBIC with the PSD-based transformation under the same noisy situation. (c) Confidence intervals for the spatially varying coefficients and the instability (over spatial points) of the potential PDEs with different support sizes.}
\label{fig:ks_result}
\end{figure}

\subsection{Spatially Dependent Kuramoto-Sivashinsky (KS) PDE}
The equation has a fourth-order derivative, making its identification very challenging because of the difficulty in obtaining an accurate estimation of the fourth-order derivative, especially in noisy situations. The equation reads as follows: 

\begin{equation}\label{eq:KS}
\begin{aligned}
u_t &= a(x) uu_x + b(x) u_{xx} + c(x) u_{xxxx};\\
a(x) = 1 + 0.25\sin(\frac{2\pi x}{20}),\, b(x) &= -1 + 0.25e^{-\frac{(x-2)^2}{5}},\, c(x) = -1 - 0.25e^{-\frac{(x+2)^2}{5}},\\
x &\in [-20, 20],\, \textrm{and} \, t \in [0, 100].
\end{aligned}
\end{equation}

\noindent The chaotic equation is numerically solved up to $t = 200$ using $N_{x} = 512$ and $N_{t} = 1024$. To limit the number of samples in our candidate library, we utilize the first half of the dataset in time (up to $t = 100$), as visualized in Figure \ref{fig:ks_dataset}. The candidate library includes powers of $u$ up to cubic, multiplied by spatial derivatives of $u$ up to fourth order. We conduct experiments with a noise level of $\epsilon = 0.01$, relatively large for the noise-sensitive solution. 

We find that the SGTR approach fails to discover the governing equation due to contaminated noise as well as the AICc loss, which unfortunately does not prefer parsimonious PDEs, as shown in Figure S2. It is worth introducing an additional penalty after applying the PSD-based transformation since it allows us to distinguish the BIC, computed with a greater complexity penalty, from the AICc, as seen in Figure \ref{fig:ks_result}(a). 

In Figure \ref{fig:ks_result}(b), despite the noisy and problematic situation, the UBIC is able to discover the governing form successfully owing to the guiding quantified PDE uncertainty, displaying a closely identical pattern that prefers the $3$-support-size (correct) parametric PDE. In Figure \ref{fig:ks_result}(c), the instability (over spatial points) is high around the Gaussian humps complicates the identification of the true symbolic structure underlying the spatially varying coefficients. More details are provided in Appendix \ref{app:symbolic_discovery}.

\section{Conclusion}
We propose a new extension of the UBIC for solving parametric PDE problems. The extended UBIC, computed with the PSD-based transformation, leverages accumulated PDE uncertainty to overcome the overfitting problem in the model selection step, successfully disambiguating the true governing parametric PDE from overfitted PDEs with dispensable candidate terms. Through extensive numerical experiments on the three canonical PDEs, we demonstrated the noise robustness and accuracy of our method in identifying the correct number of effective terms and their varying coefficients, which represent the parametric dependencies of the governing PDE. Our results suggest that the extended UBIC is a powerful criterion for parametric PDE discovery, emphasizing the parsimony of governing equations. The ability to compute confidence intervals for varying coefficients further enhances the interpretability of potential models, providing comprehensive insights into their stability. This work not only advances the methodology for PDE discovery but also establishes a fundamental framework applicable to a wide range of complex dynamical systems. To improve the practicality of our method, we plan to use the proposed UBIC as the fitness function in a genetic algorithm based PDE discovery framework, freeing ourselves from the overcompleteness assumption on the initialized candidate library.

\bibliography{acml24}

\appendix
\begin{table}[ht]
    \centering
    \caption{Symbolic expression of varying coefficients}
    \begin{tabular}{|c|c|c|c|}
        \hline
        Dataset & Noise level ($\epsilon$) & Discovered varying coefficients & \%Coefficient error\\
        \hline
        Burgers & \makecell[c]{$2$\\$4$} & \makecell[c]{$0.09441, -0.98643-0.25067\sin(t)$\\$0.08080, -0.25146\sin(t)-0.95987$} & \makecell[c]{$1.286, 5.591$\\$19.20, 3.813$}\\
        \hline
        AD  & \makecell[c]{$2$\\$4$} & \makecell[c]{$-\sin(1.2458x), \cos(1.2499x)-1.4502, 0.08736$\\$-\sin(1.2415x), \cos(1.244x)-1.4216, 0.06406$} & \makecell[c]{$20.05, 2.958, 12.64$\\$20.06, 4.554, 35.94$}\\
        \hline
        KS & $0.01$ & \makecell[c]{$0.25113\sin(0.32253x) + 0.95955$\\$-0.976754 + 0.353986e^{- 0.493521(1 - 0.547288x)^{2}}$\\$-0.966723 - 0.249388e^{- 3.73627(0.429437x + 1)^{2}}$} & \makecell[c]{$4.055$\\$4.027$\\$4.365$}\\
        \hline
    \end{tabular}
    \label{tab:ce}
\end{table}

\section{Symbolic discovery of varying coefficients} \label{app:symbolic_discovery}
Symbolic discovery of varying coefficients is achieved via the PySR package \citep{PySR}. To prioritize parsimonious expressions of varying coefficients, we consider model rankings based on the PySR's score. We evaluate any selected interpretable expression $\hat{h}(x)$ against its ground truth $h(x)$ using the percentage relative coefficient error: $\mathrm{CE}(h(x), \hat{h}(x)) = 100\times\frac{\norm{\hat{h}(x)-h(x)}_{1}}{\norm{h(x)}_{1}}$. We note that $\mathrm{CE}(h(t), \hat{h}(t))$ are calculated in the same manner. Table \ref{tab:ce} lists the relative coefficient errors for every experiment in this paper. In the experimental cases of the parametric Burgers' and AD PDEs, we can uncover the correct mathematical expressions/structures of the varying coefficients with acceptable accuracy. For the parametric KS PDE case, we cannot initially retrieve the correct expression only for the varying coefficient of $u_{xxxx}$, as the suggested expression by PySR is $-0.9798 + 0.19909xe^{-0.26832x^{2}}$, which however hints that a common Gaussian function should be used instead due to its similar accuracy. To refine the initial expressions for $u_{xx}$ and $u_{xxxx}$, we use Feyn's autorun functionality \citep{Feyn} with minimal complexity settings, ultimately discovering the Gaussian formulas. The coefficient errors are satisfactory, less than $5$\%. 

\begin{figure}[t]
\centering
\begin{subfigure}
	\centering
	\includegraphics[width=0.8\textwidth]{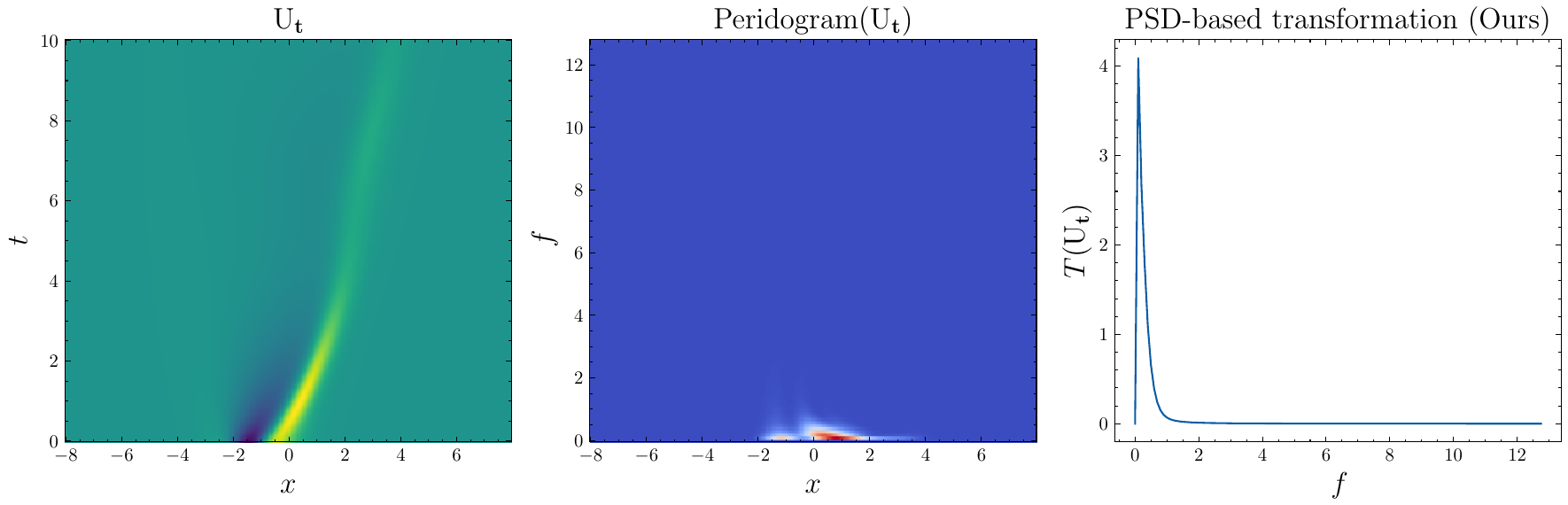}
	\caption*{$\epsilon = 0$ (clean)}
\end{subfigure}
\begin{subfigure}
	\centering
	\includegraphics[width=0.8\textwidth]{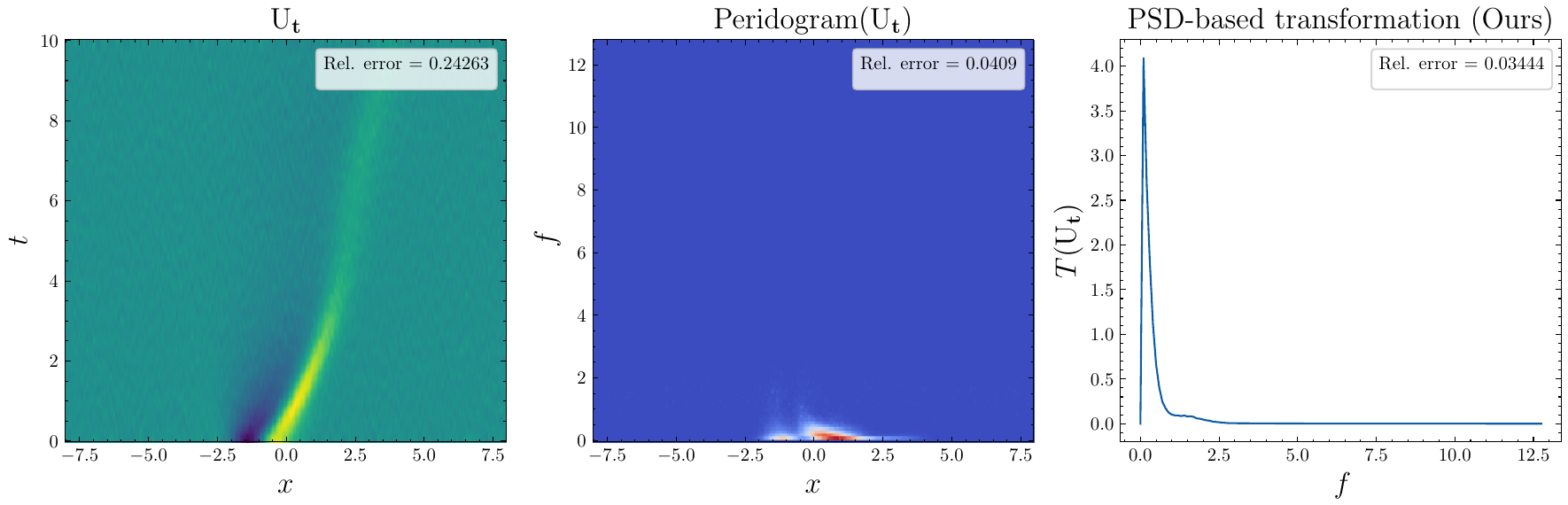}
	\caption*{$\epsilon = 4$ (with Gaussian noise)}
\end{subfigure}
\caption{\textbf{Parametric Burgers' PDE.} We visualize different representations of $\vec{\mathrm{U}_{t}}$ and measure their relative errors against the ground truth data.}
\label{fig:burgers_psd}
\end{figure}

\section{Noise-robustness of PSD} \label{app:psd}
We explore the noise-robustness of our PSD-based transformation using the Burgers' PDE as an example. In Figure \ref{fig:burgers_psd}, different representations of $\vec{\mathrm{U}_{t}}$ are presented. We evaluate the accuracy of each representation by comparing it to the ground truth data using the relative Frobenius-norm error. The fact that our PSD-based transformed representation matches its ground truth more closely than other representations under the noisy situation demonstrates its superior robustness. Therefore, we apply our extended UBIC with the PSD-based transformation. 

\end{document}